\documentclass[runningheads]{llncs}
\usepackage{graphicx}
\usepackage{tikz}
\usepackage{comment} 
\usepackage{amsmath,amssymb} % define this before the line numbering.
\usepackage{color}
\usepackage{cite}
\usepackage{caption}
\usepackage{subcaption}

\begin{document}
% \renewcommand\thelinenumber{\color[rgb]{0.2,0.5,0.8}\normalfont\sffamily\scriptsize\arabic{linenumber}\color[rgb]{0,0,0}}
% \renewcommand\makeLineNumber {\hss\thelinenumber\ \hspace{6mm} \rlap{\hskip\textwidth\ \hspace{6.5mm}\thelinenumber}}
% \linenumbers
\pagestyle{headings}
\mainmatter
\def\ECCVSubNumber{***}  % Insert your submission number here

\title{Blind Motion Deblurring through SinGAN Architecture} % Replace with your title

% INITIAL SUBMISSION 
\begin{comment}
\titlerunning{ECCV-20 submission ID \ECCVSubNumber} 
\authorrunning{ECCV-20 submission ID \ECCVSubNumber} 
\author{Anonymous ECCV submission}
\institute{Paper ID \ECCVSubNumber}
\end{comment}
%******************

% CAMERA READY SUBMISSION
%\begin{comment}
\titlerunning{Blind Motion Deblurring through SinGAN}
% If the paper title is too long for the running head, you can set
% an abbreviated paper title here
%
\author{Harshil Jain \and Rohit Patil \and Indra Deep Mastan \and Shanmuganathan Raman}

%
%\authorrunning{F. Author et al.}
% First names are abbreviated in the running head.
% If there are more than two authors, 'et al.' is used.
%
\institute{Computer Science and Engineering, \\Indian Institute of Technology Gandhinagar.}

%\end{comment}
%******************
\maketitle

\begin{abstract}
Blind motion deblurring involves reconstructing a sharp image from an observation that is blurry. It is a problem that is ill-posed and lies in the categories of image restoration problems. The training data-based methods for image deblurring mostly involve training models that take a lot of time  \cite{kupyn2018deblurgan, ramakrishnan2017deep}. These models are data-hungry \textit{i.e.}, they require a lot of training data to generate satisfactory results. Recently, there are various image feature learning methods developed which relieve us of the need for training data and perform image restoration and image synthesis, e.g., DIP \cite{Ulyanov2018CVPR}, InGAN \cite{shocher2018internal}, and SinGAN \cite{shaham2019singan}. SinGAN is a generative model that is unconditional and could be learned from a single natural image. This model primarily captures the internal distribution of the patches which are present in the image and is capable of generating samples of varied diversity while preserving the visual content of the image. Images generated from the model are very much like real natural images. In this paper, we focus on blind motion deblurring through SinGAN architecture. 
\end{abstract}

%In contrast to previous single image GAN frameworks, SinGAN is not conditional and limited to texture images. 
\section{Introduction}
Image deblurring is the task to restore the images where the corruption occurs due to blurring; for example, motion blur due to camera shake. It could also be used as a preprocessing step for other problems. For example, real-time object identification is better achieved when prepossessing the input image through deblurring. There are several methods proposed for solving the image deblurring problem using Generative Adversarial Net (GAN) \cite{kupyn2018deblurgan, zheng2019edge}, which have the capability to restore image features that are highly corrupted due to blur. They could model the non-uniform image blur arising from the input images. They are mostly studies in the conditional GAN  setting where deblurring is performed conditioned on the edges and other features extracted from the images.

The motion image deblurring methods primarily involve extensive training on a large dataset of images (e.g., \cite{kupyn2018deblurgan,ramakrishnan2017deep}). Hence, the main requirement for training these deep models is high-end graphic processing units (GPUs) and a lot of resources owing to extensive computational power. Researchers who do not have access to these resources cannot carry out extensive research. Hence, our main motivation is to train the model on a single image, thus reducing the computational power and cost.

There is a vast interest in developing deep features learning frameworks that do not use the training data. These single image deep feature learning methods have shown remarkable performance for image restoration and image synthesis tasks \cite{Ulyanov2018CVPR, shocher2018zero, mastan2019multi, shocher2018internal, shaham2019singan, sidorov2019deep, deep2019dcil, ren2020neural}. Ulyanov \textit{et al.} showed a handcrafted structure of the convolution neural network (CNN) works as the implicit prior to solving the image restoration, named deep image prior (DIP) \cite{ulyanov2018deep}. Mataev \textit{et al.} proposed DeepRED, which leveraged DIP by using explicit regularization called RED (regularization by denoising) \cite{mataev2019deepred}. DeepRED is shown to work for the Uniform filter and Gaussian filter deblurring.  SelfDeblur performs self-supervised learning for image deblurring \cite{ren2020neural}.  InGAN \cite{shocher2018internal} and SinGAN \cite{shaham2019singan}  architectures are single image GAN frameworks that perform internal learning with no prior and works well for image synthesis tasks. Inspired by the SinGAN \cite{shaham2019singan}, this work is closely related to DeblurGAN \cite{kupyn2018deblurgan} and SinGAN\cite{shaham2019singan}.

\section{Background}
In this section, we first define the image deblurring problem. Then we discuss training data-based adversarial framework DeblurGAN \cite{kupyn2018deblurgan} and single image GAN framework SinGAN \cite{shaham2019singan}. 
\subsection{Image Deblurring}
The image and videos that are captured by devices such as mobile phones, cameras mostly contain motion blur artifacts, which are caused by a variety of factors, including camera shake and scene content, which is dynamic in nature. If the exposure time is fixed, then the sensor averages the signal across the points in the scene for any movement during the recording. The reconstruction of the sharp latent image from an observation that is blurry is a problem that is not posed correctly, and it can be categorized into two categories - blind or non-blind deconvolution depending on whether we have information regarding the camera shake or not. One way to formulate image deblurring is to assume the possible sources that can cause the blur in the image. One could also reduce blur caused by camera shake by assuming uniform blur across the image.

One can formulate a non-uniform blur kernel as follows:
\begin{equation}
I_{Blur} = kernel(M) * I_{Sharp} + N
\end{equation} 
where $I_{Blur}$ is the blurred image, $k(M)$ are blur kernels which are unknown and are determined by motion filed $M$. $I_{Sharp}$ is the sharp latent image, $*$ denotes the convolution, and N is additive noise. Non-blind deblurring is based on the assumption that the blur kernels $kernel(M)$ are known. However, for blind deblurring, given the blurred image $I_B$, the goal is to determine the sharp image $I_{Sharp}$ and blur kernels $kernel(M)$.

\subsection{DeblurGAN}
DeblurGAN \cite{kupyn2018deblurgan} is an end-to-end learned method for motion deblurring. DeblurGAN achieves good results for structural similarity and visual appearance. The concise version explaining all details is as follows.

\subsubsection{Model Overview.}
DeblurGAN is a conditional GAN with a modified loss function (multi-content). In a conditional GAN, both the generator and the discriminator models depend upon the class label. When the generator tries to generate images of its own, an additional input (label or class) is given to it to generate images corresponding to the label.
 
The generator $G_{\theta_{G}}$ is mainly responsible for deblurring.  $I_{Blur}$ is the input to the generator, and $I_{Sharp}$ is the corresponding output. The discriminator $D_{\theta_{D}}$ is responsible for distinguishing the generated image $I_{Sharp}$ and the sharp latent image. Thus, both the networks are trained in an adversarial manner.

\subsubsection{Loss Function.} It is combination of adversarial loss $L_{GAN}$ and VGG feature loss $L_{vgg}$, \textit{i.e.}, $L = L_{GAN} + \lambda L_{vgg}$. $L_{GAN}$ and $L_{vgg}$ are defined as follows.

\begin{equation}
L_{GAN} = \sum\limits_{n=1}^N -D_\theta (G_{\theta_G} (I_{Blur})) 
\end{equation}

\begin{equation}
L_{vgg} = \frac{1}{W_{i,j}H_{i,j}}\sum\limits_{x=1}^{W_{i,j}}\sum\limits_{y=1}^{H_{i,j}}(\phi_{i,j}(I_{Sharp})_{x,y}-\phi_{i,j}(G_{\theta_G} (I_{Blur}))_{x,y})^2.
\end{equation}
Here, $\phi$ denotes VGG19 network pretrained on ImageNet dataset. $\phi_{i,j}$ is the feature map which is obtained by the activation of the jth convolution layer before the ith max-pooling layer in $\phi$. $W_{i,j}$ and $H_{i,j}$ are the height and width of the features maps $\phi_{i,j}$.

\subsubsection{Training Details.} The generator network takes the blurred image as input and outputs the restored image. The discriminator takes the restored and sharp images and calculates the distance between them. The overall loss consists of the discriminator's WGAN loss and the perceptional loss.

\subsection{SinGAN}
SinGAN is an unconditional generative model, learning from just one image. This model primarily captures the internal distribution of different patches within the image at different scales and then produce samples of varied variety and retain the visual quality of the image. In contrast to previous single image GAN frameworks, SinGAN is not conditional. Images generated from the model are very much like real natural images.

\subsubsection{Why to use SinGAN for deblurring?}
 SinGAN is very robust in dealing with general natural images containing textures and structures which are complex without relying on the presence of the images of the same class in the training data. The primary reason for this is the architecture, which consists of a pyramid of fully convolutional GANs where the task of each generative network is to capture the patch distribution at a specific different scale. SinGAN is successful in producing images that are semantically similar to the training image apart from enhancing the image with the overall placement of objects. SinGAN has produced good quality results for image super-resolution. Thus, we have taken a step further to explore SinGAN for image deblurring.

\section{Blind Motion Deblurring through SinGAN}
\begin{figure}
\includegraphics[width=\textwidth]{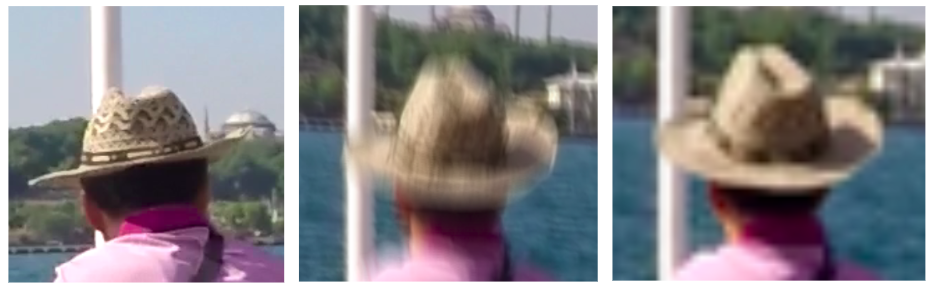}\\
(a) Training Image (sharp) \hspace{0.4cm} (b) Input blurry image \hspace{1cm} (c) Restored Image 
\caption{The model was trained on the leftmost image, the inference was made on the image in the center (at a slightly different angle than the training image), and the rightmost image is the  image generated by the model.}
\label{fig: results1}
\end{figure}

The main idea is first to train the SinGAN model using a \textbf{single sharp image} and then perform the image deblurring of the blurred image, which is similar to the sharp image but taken at a different angle. We believe that the approach simulates the real-world scenario where one capture multiple pictures, but unfortunately, one of the images turns out to be blurry due to camera shake. In Fig.~\ref{fig: results1}, illustrate the image deblurring results of our approach. 

\subsection{Architecture.} 
\begin{figure}[!h]\centering 
       \includegraphics[scale=0.65]{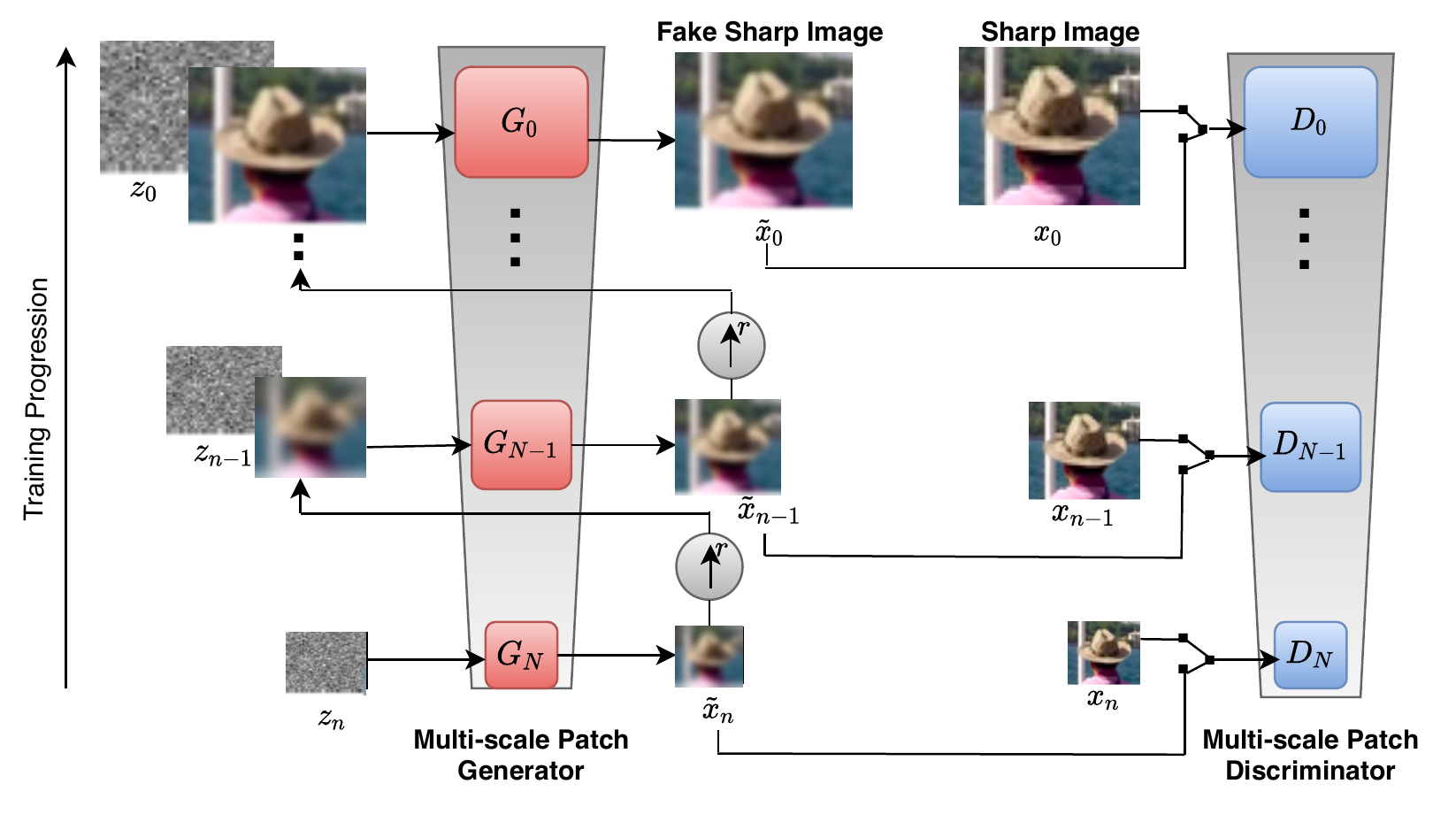}
       \caption{The figure illustrates SinGAN's multi scale pipeline for image deblurring.}
       \label{fig: singanMultiscale}
\end{figure}

Our proposed model for blind motion deblurring is based on SinGAN architecture. It consists of $n$ GANs stacked on top of one another, where each GAN is responsible for generating an image at a different scale. The training is done in such a way that $G_N$ is responsible for capturing coarse features of the image, while $G_0$ captures the finer features of the image. At the time of inference, the finest scale GAN is used iteratively. At each scale, $G_N$ is responsible for generating the image samples that contains patches that are indistinguishable for the from the patches in the down-sampled training image, $x_n$, by the discriminator $D_N$. The effective patch size decreases as we traverse from the bottom to top in the network. The input to $G_N$ is a random noise image $z_n$, and the image which is generated from the previous scale $\tilde{x}_n$ that is upsampled to the current resolution. This does not apply to the coarsest level, which is only generative in nature. The generation process at level $n$ involves all generators \{$G_N,..., G_n$\} and all noise maps \{$z_N, ..., z_n$\} up to this level. A diagrammatic representation of the entire architecture is as shown in Fig.~\ref{fig: singanMultiscale}.

\begin{figure}[!h]\centering 
       \includegraphics[scale=0.45]{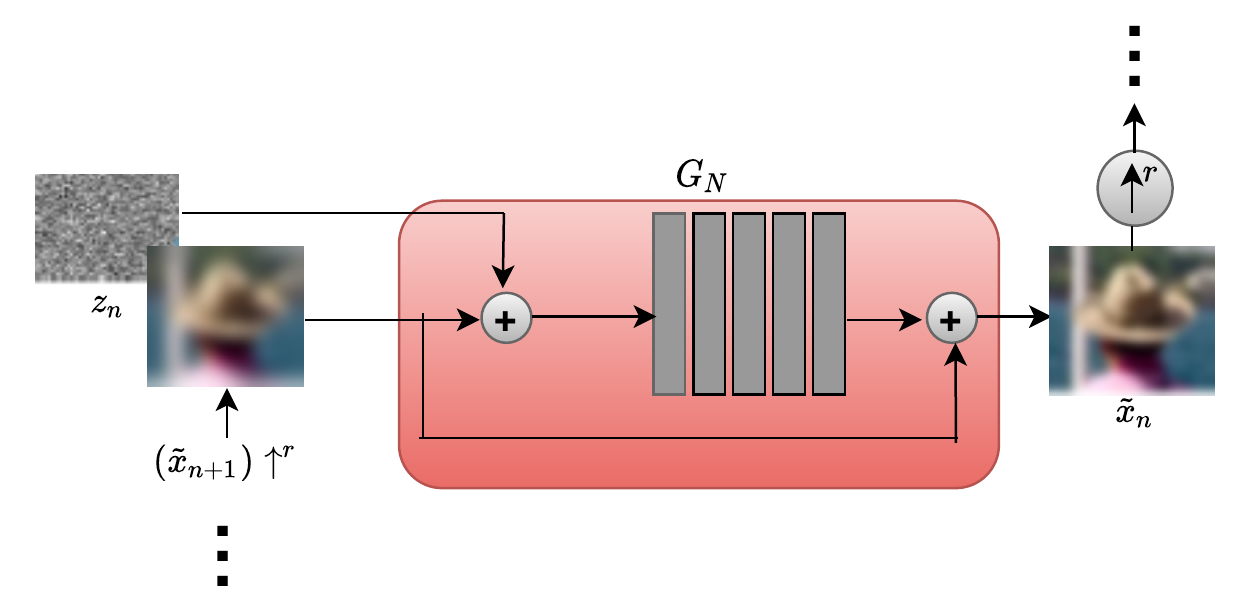}
\caption{The figure illustrates SinGAN's single scale generation.}
       \label{fig: singanSinglescale}
\end{figure}

The image features generation at a single scale is shown in Fig.~\ref{fig: singanSinglescale}. At each scale $n$, the image from the previous scale, ${\tilde{x}_{n+1}}$, is upsampled and added to the input noise map, $z_n$. The result is passed through five convolution layers that output a residual image, which is added back to (${\tilde{x}_{n+1}}$) $\uparrow^r$. It is the output ${\tilde{x}_{n}}$ of the generative network $G_N$.

\subsubsection{Loss Function.}
The loss function is a combination of an adversarial loss term, $L_{adv}$ and a reconstruction loss term, $L_{rec}$. We describe the objective function for the training for the $n^{th}$ GAN in Eq.~\ref{eq: objective}.  
\begin{equation}\label{eq: objective}
\min_{G_N}\max_{D_N} L_{adv}(G_N,D_N) + \alpha L_{rec}(G_N)
\end{equation}
The model is trained in a sequential fashion from the coarsest scale for which $G_N$ is responsible to the finest one which is handled by $G_0$. The adversarial loss and reconstruction loss are described in detail in SinGAN \cite{shaham2019singan}. The reconstruction loss is given in Eq.~\ref{eq: rec} for completeness. 
\begin{equation}\label{eq: rec}
L_{rec} = ||G_N(0,({\tilde{x}_{n+1}^{rec}}\uparrow^{r})-x_n||
\end{equation}
The adversarial loss $L_{adv}$ puts a penalty on the distance between the patch distribution in $x_n$ and the patch distribution in the generated samples ${\tilde{x}_{n}}$. The reconstruction loss $L_{rec}$ ensures for a principle features in the manipulation of images which is the existence of noise maps that are responsible for producing $x_n$.

\subsection{Details of Architecture Used at Inference}
\begin{figure}\centering 
	\includegraphics[scale=0.75]{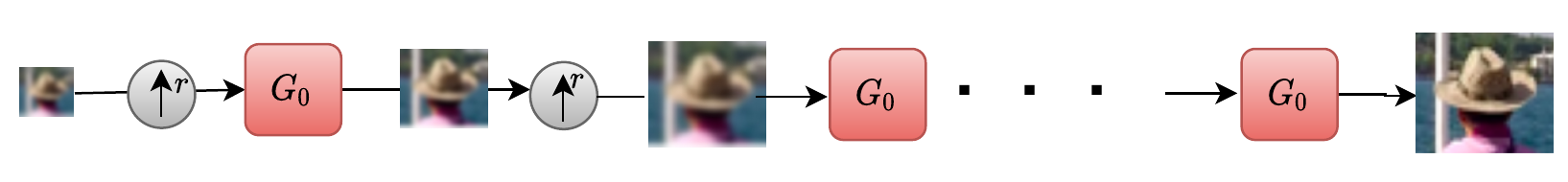}
	\caption{The figure illustrating the proposed model for inference}
	\label{fig: singantest}
\end{figure}

We train the multi-scale model on the sharp ground truth image. We use the weights obtained after the entire model is trained on a single sharp image. At test time, we iteratively upsample the deblurred image and pass it through the GAN which extracts fine features, $G_0$. A diagrammatic representation of the architecture used at the time of inference is given in  Fig.~\ref{fig: singantest}.

\subsection{Training Details}
The sharp ground truth image is trained up to eight scales as we traverse from the bottom to top in the pyramid, that is from the coarsest scale to the finest scale. 
While inferencing, we are using the weights obtained after training and upsampling iteratively and feeding it to the finest level generator. In what follows, we provide visual results for image samples taken from GoPro dataset.

\section{Results}
\begin{figure}
	\includegraphics[width=\textwidth,height=6.5cm]{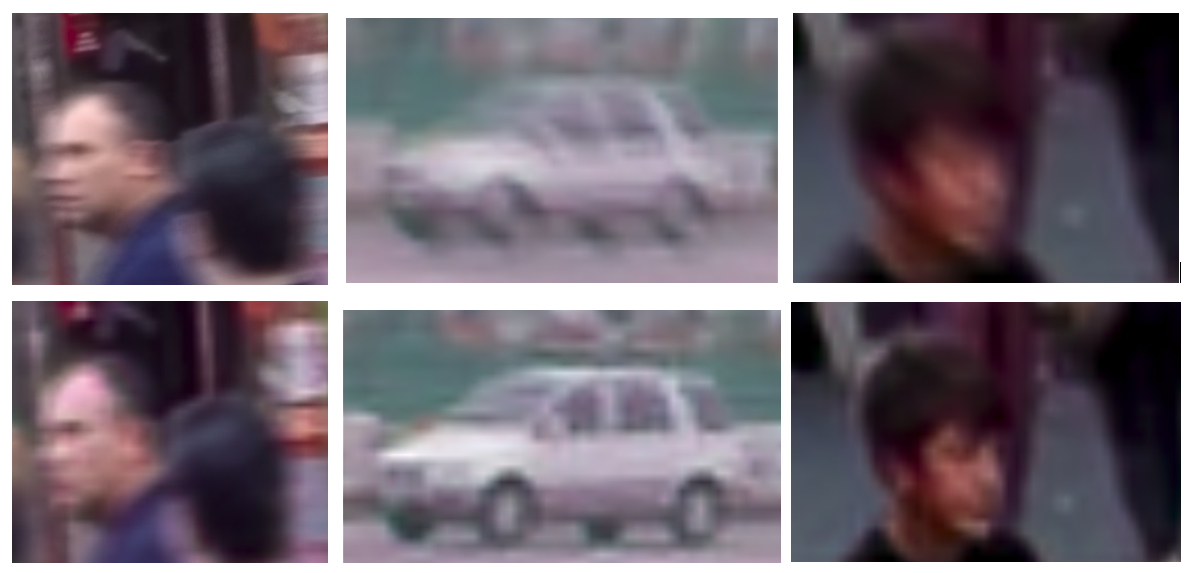}
	\caption{The figure shows the results produced by our apprach. The images in first row (top) are the input blurry images and images in second row (bottom) are the restored images.}
	\label{fig: results2}
\end{figure}

We illustrated image deblurring results in Fig.~\ref{fig: results2}. The metrics used for measuring accuracy are SSIM and PSNR. SSIM stands for Structural Similarity Index and is a value between -1 and 1; -1 being that two images are very different from each other, and value equal to 1 means that the two images are very close to each other. PSNR stands for Peak to Signal Noise Ratio and is the mean squared error between the original image and the reconstructed image. The SSIM on a sample of around 100 images randomly chosen from the GoPro Test Dataset is found to be 0.52. The PSNR on a sample of around 100 images randomly chosen from the GoPro Test Dataset is found to be 15.3. 

\section{Conclusion}
We show how to perform training data-independent blind motion deblurring using SinGAN. The  training data-independent solution to the blind motion deblurring problem could generalize well by learning features from a single image. The challenge comes when the features present in the image are not enough to perform the restoration. It is interesting to note that SinGAN has reported similar limitations for image editing tasks.

\section{Acknowledgement}
We thank Aalok Gangopadhyay for useful discussions and comments at the beginning of the project.

% ---- Bibliography ----
%
% BibTeX users should specify bibliography style 'splncs04'.
% References will then be sorted and formatted in the correct style.
%
\bibliographystyle{splncs04}
\bibliography{egbib}

\begin{thebibliography}{10}
\providecommand{\url}[1]{\texttt{#1}}
\providecommand{\urlprefix}{URL }
\providecommand{\doi}[1]{https://doi.org/#1}

\bibitem{mastan2019multi}
Deep~Mastan, I., Raman, S.: Multi-level encoder-decoder architectures for image
  restoration. In: Proceedings of the IEEE Conference on Computer Vision and
  Pattern Recognition Workshops. pp.~0--0 (2019)

\bibitem{deep2019dcil}
Deep~Mastan, I., Raman, S.: Dcil: Deep contextual internal learning for image
  restoration and image retargeting. WACV  (2020)

\bibitem{kupyn2018deblurgan}
Kupyn, O., Budzan, V., Mykhailych, M., Mishkin, D., Matas, J.: Deblurgan: Blind
  motion deblurring using conditional adversarial networks. In: Proceedings of
  the IEEE Conference on Computer Vision and Pattern Recognition. pp.
  8183--8192 (2018)

\bibitem{mataev2019deepred}
Mataev, G., Milanfar, P., Elad, M.: Deepred: Deep image prior powered by red.
  In: Proceedings of the IEEE International Conference on Computer Vision
  Workshops. pp.~0--0 (2019)

\bibitem{ramakrishnan2017deep}
Ramakrishnan, S., Pachori, S., Gangopadhyay, A., Raman, S.: Deep generative
  filter for motion deblurring. In: Proceedings of the IEEE International
  Conference on Computer Vision. pp. 2993--3000 (2017)

\bibitem{ren2020neural}
Ren, D., Zhang, K., Wang, Q., Hu, Q., Zuo, W.: Neural blind deconvolution using
  deep priors. In: Proceedings of the IEEE/CVF Conference on Computer Vision
  and Pattern Recognition. pp. 3341--3350 (2020)

\bibitem{shaham2019singan}
Shaham, T.R., Dekel, T., Michaeli, T.: Singan: Learning a generative model from
  a single natural image. In: Proceedings of the IEEE International Conference
  on Computer Vision. pp. 4570--4580 (2019)

\bibitem{shocher2018internal}
Shocher, A., Bagon, S., Isola, P., Irani, M.: Internal distribution matching
  for natural image retargeting. arXiv preprint arXiv:1812.00231  (2018)

\bibitem{shocher2018zero}
Shocher, A., Cohen, N., Irani, M.: “zero-shot” super-resolution using deep
  internal learning. In: Proceedings of the IEEE Conference on Computer Vision
  and Pattern Recognition. pp. 3118--3126 (2018)

\bibitem{sidorov2019deep}
Sidorov, O., Yngve~Hardeberg, J.: Deep hyperspectral prior: Single-image
  denoising, inpainting, super-resolution. In: Proceedings of the IEEE
  International Conference on Computer Vision Workshops. pp.~0--0 (2019)

\bibitem{Ulyanov2018CVPR}
Ulyanov, D., Vedaldi, A., Lempitsky, V.: Deep image prior. In: The IEEE
  Conference on Computer Vision and Pattern Recognition (CVPR) (June 2018)

\bibitem{ulyanov2018deep}
Ulyanov, D., Vedaldi, A., Lempitsky, V.: Deep image prior. In: Proceedings of
  the IEEE Conference on Computer Vision and Pattern Recognition. pp.
  9446--9454 (2018)

\bibitem{zheng2019edge}
Zheng, S., Zhu, Z., Cheng, J., Guo, Y., Zhao, Y.: Edge heuristic gan for
  non-uniform blind deblurring. IEEE Signal Processing Letters
  \textbf{26}(10),  1546--1550 (2019)

\end{thebibliography}
\end{document}